\documentclass[letterpaper, 10 pt, conference]{ieeeconf}  

\IEEEoverridecommandlockouts                              

\usepackage{graphics} 
\usepackage[pdftex]{graphicx}
\graphicspath{{./Figures/}}
\DeclareGraphicsExtensions{.pdf,.jpeg,.png}
\usepackage{epsfig} 
\usepackage{amsmath} 
\usepackage{amssymb}  
\usepackage{bbold}
\usepackage[export]{adjustbox}
\usepackage[]{subfig}
\usepackage{lipsum}
\usepackage{algorithm}%
\usepackage{paralist}

\usepackage{bm}
\newcommand{\vg}[1]{\bm{#1}}
\renewcommand{\v}[1]{\mathbf{#1}}
\usepackage{algpseudocode}
\usepackage{color}
\DeclareSymbolFont{matha}{OML}{txmi}{m}{it}
\DeclareMathSymbol{\varv}{\mathord}{matha}{118}

\title{\LARGE \bf
Design-Informed Kinematic Control for Improved Dexterous Teleoperation of a
Bilateral Manipulator System }
\author{Lasitha Wijayarathne, Juan Vallejo, Anthony Barnum, Zachary Cloutier \\  and Frank L. Hammond III, \textit{IEEE Member}
\thanks{Lasitha Wijayarathne, Anthony Barnum, Zachary Cloutier, Juan Vallejo and Frank Hammond are with the Woodruff School of Mechanical Engineering,
        Georgia Institute of Technology, Atlanta, GA, USA.}
\thanks{Frank L. Hammond III is with the Woodruff School of Mechanial Engineering and the Coulter Department of Biomedical Engineering at the Georgia Institute of Technology. {\tt\small frank.hammond@me.gatech.edu.}}%
}

\begin{document}
\maketitle
\thispagestyle{empty}
\pagestyle{empty}

\begin{abstract}
This paper explores the possibility of improving bilateral robot manipulation task performance through optimizing the robot morphology and configuration of the system through motion. To optimize the design for different scenarios, we select a set of tasks that represent the variability in small scale manipulation (e.g. pick and place, tasks involving positioning and orientation) and track the motion to obtain a reproducible trajectory. Kinematic data is captured through an electromagnetic (EM) tracker system while a human subject performs the tasks. Then, the data is pre-processed and used to optimize the morphology of each symmetric robot arm of the bilateral system. Once optimized, a kinematic control scheme is used to generate a motion with dexterous configurations. The dexterity is evaluated along the trajectories with standard dexterity metrics. Results show a 10\% improvement in dexterous maneuverability with the optimized arm design and optimal base configuration.
\end{abstract}

\section{Introduction}
\begin{figure}[t!] 
\centering
\includegraphics[width=0.78\linewidth]{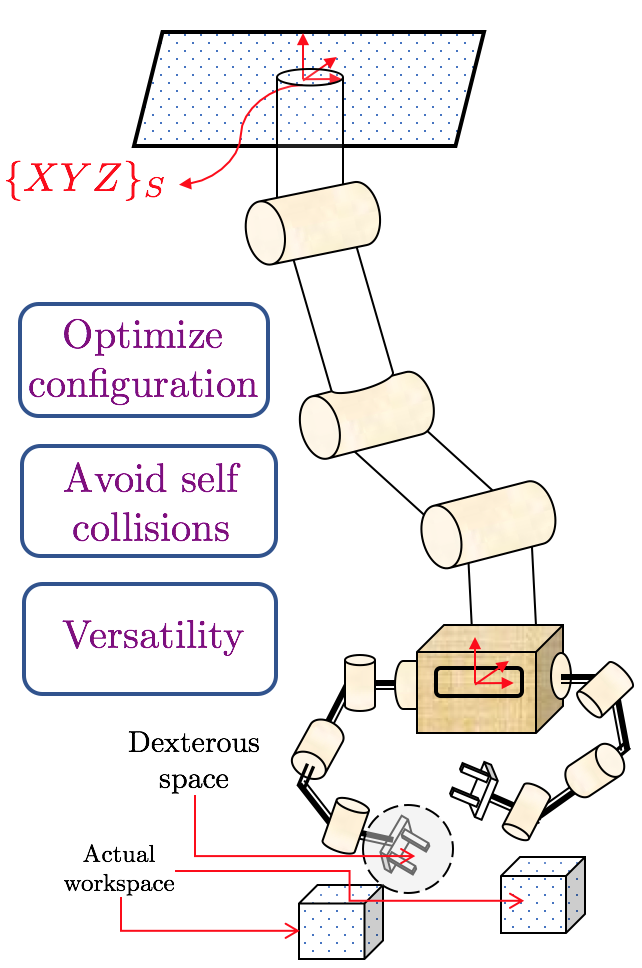}
\caption{Illustration of proposed bilateral manipulator system}
\label{fig:setup_intro}
\vspace{-15pt}
\end{figure}

Execution of manipulation tasks in the operational (or task) space is commonplace nowadays. This strategy applies to both kinematic and force-torque control systems.  If the robotic system is kinematically redundant for the given task, it is possible to perform additional secondary tasks simultaneously, as exemplified in Figure \ref{fig:setup_intro}. A general approach is to execute them in a prioritized or combined manner using null space projections. These secondary tasks could include obstacle avoidance, visibility improvement, and configuration optimization using the current design morphology.\par

Dexterous robotic manipulation has been long studied, and many commercial serial and parallel manipulators are available for specialized and general applications. Among those, redundant manipulators are more appealing since they have a well-connected work-space and improved dexterity. Multiple arm manipulator systems (kinematic trees) take a vital place in terms of redundancy since they can be utilized to enhance the performance for a given task. The performance of a robot system inherently relies on its geometrical design and control scheme. Underlying control schemes could be used to improve the performance but are limited by the bounds of the design. Thus, improved design can help enhance the performance of the overall robot system.
\par
\section{Motivation}
Today's robots are good at performing specialized tasks \cite{Dietrich2020HierarchicalRobots} but not on everyday tasks that involve shifting workspaces, require adaptation and dexterity. While bilateral manipulation delivers more combined manipulability to coordinated tasks encountered on a daily basis \cite{Smith2012DualSurvey, Park2016Dual-armEvaluation}, we still do not leverage the knowledge of the design domain in motion generation. 
\par In this paper, we layout a formulation to optimize the design morphology of a bilateral system and use it in the motion generation of the entire system at a kinematic level. Our main contributions are:
\begin{itemize}
    \item \textbf{A method for a data driven optimization of a bilateral robot system morphology}
    \item \textbf{Design informed optimization of the global positioning of the bilateral system to achieve improved dexterity}
\end{itemize}

\section{Related Work}
In the literature, design optimization of serial/parallel robot manipulators has been widely studied. These works can be mainly categorized into two sub-categories: design optimization for general and specific tasks. Studies of \cite{Paden2010DesignManipulators, Ranjbaran1995TheIsotropy}, \cite{Vijaykumar1986} can be recognized as attempts to optimize the design for general-tasks. In these, non-redundant manipulators have been considered where analysis is primarily done analytically. On the other hand, works of  \cite{Rodriguez2013EffectorActuation,Shirafuji2019KinematicKinematicsb,Shiakolas2002,Rao1989OptimizationSurwef} can be regarded as optimizations for specific tasks. Tasks in these vary from medical applications to construction work tasks and non-redundant, redundant, and continuum manipulators have been considered. 
Moreover, \cite{Webster2018KinematicPlanning} used RRT (rapidly exploring random trees) to optimize a continuum robot design for motion optimality and visibility. Here, the design is optimized for motion and design. \par

Design optimization is broad in scope and the formulation used for optimization is important to ensure the design is robust and physically realizable. Since the robot's kinematic model and desired cost function are non-linear, most of the cited work here use global optimization techniques. Simulation annealing (SA) and genetic algorithms (GA) are widely used as global optimization techniques. While the use of stochastic methods inhibits the reproducibility of results, it could guarantee that the optimal parameters are found (in a sense of probabilistic completeness\cite{Webster2018KinematicPlanning}. In addition to design, optimizing for motion simultaneously has gained attention. \cite{Computational-Co-Optimization-of-Design-Parameters-and-Motion-Trajectories-for-Robotic-Systems-Paper.pdf} use a co-optimization framework to iteratively optimize for motion and design parameters simultaneously. This method is well suited for a design with good prior morphology (with a fewer DOFs) as this method is a local optimization technique. \par

Kinematic and dynamic metrics are considered as primary or secondary metrics used for design candidate assessment. Studies of 
 \cite{Gosselin1991AManipulators, Hammond2009MorphologicalMeasures},  \cite{Asada2009ADesign, KhatibOptimizationManipulators} use kinematic and dynamic isotropic metrics for evaluation, and variants of isotropic measures (e.g., weighted isotrophy, dynamic isotrophy) are discussed in detail. In contrast, designs for specific tasks are evaluated with cost functions built using desired kinematic profiles \cite{Shirafuji2019KinematicKinematics}.

Moreover, all the above work except for \cite{Paden2010DesignManipulators, Shirafuji2019KinematicKinematics} use the DH convention to represent the manipulators. DH (Denavit-Hartenberg) representation is suited in cases where orientation is not taken into account. If the objective function consists of the full pose (position+orientation), a representation of orientation should be used (e.g. quaternion, euler) where distance metrics of position and orientations are not matched. This would result in poor inverse differential solutions compared to the manifold approach used in the product of exponential (PoE) method.

\par
Existing bilateral systems similar to the proposed system in this paper can be cited as 
\cite{Hannaford2013Raven-II:Research, GuthartTheApplication}. These systems are built primarily for surgical tele-operation and are equipped with a fixed base. Furthermore, the fundamentals of autonomous bi-manual tasks have been studied in these works \cite{Park2016Dual-armEvaluation}, \cite{Shiakolas2002}
\cite{Lee2015RedundancyExperimentsb}. The work presented in this paper could be extended easily to autonomous tasks with a focus on dexterity and configuration optimization.

\section{Fundamentals and Preliminaries}
\label{sec:fundamentals}
\begin{figure}[!t] 
\centering
\includegraphics[width=0.9\linewidth]{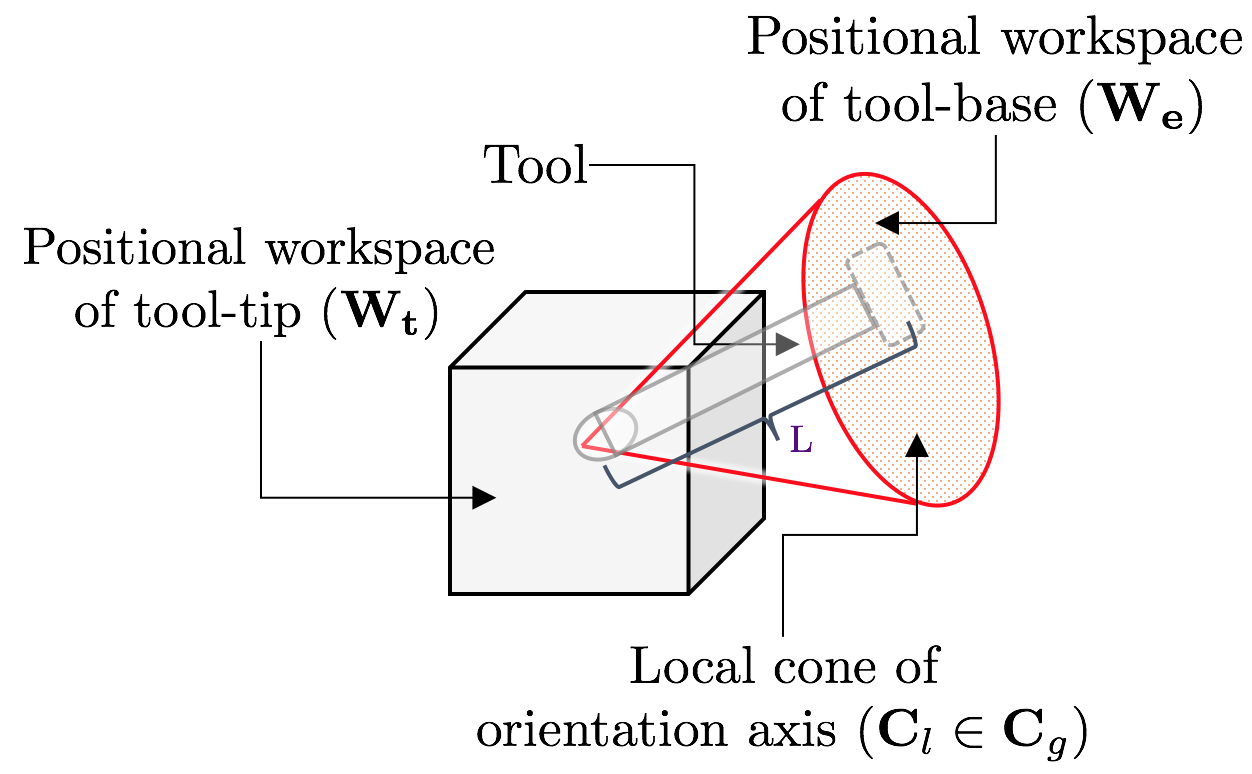} 
\caption{Illustration of the relation of work-space and the projection of it on the base of the tool}
\label{fig:tool_setup}
\vspace{-16pt}
\end{figure}
\subsection{Small Scale Bi-manual Tasks}
Small scale bi-manual tasks are found in many application domains (e.g., medical domain, micro-assembly, and biological tissue manipulation). \textbf{Compared to large scale manipulation tasks (\cite{Rakita2019SharedManipulation, Park2016Dual-armEvaluation}, small scale manipulation tasks generally require additional dexterity over workspace volume}. Moreover, the capability to generate external wrenches in desired directions is essential in small scale manipulation tasks\cite{Wijayarathne2017}. 
\par
\begin{figure*}[t!] 
\centering
\includegraphics[width=\textwidth,height=5.0cm]{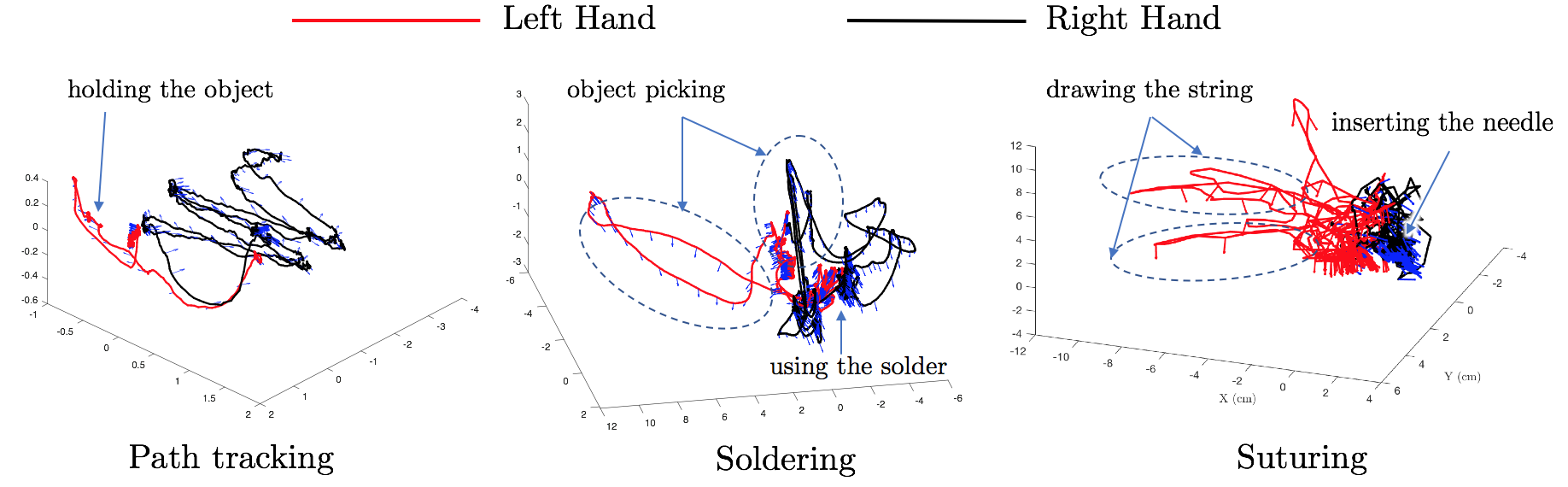}
\caption{Task kinematic profiles. Motion primitives of small tasks. Characteristics include High Dynamic Bandwidth, Positional and Orientation Dexterity. Red: Left Arm, Black: Right Arm}
\label{fig:EM_data}
\vspace{-10pt}
\end{figure*}

This paper focuses on using kinematic data from human subject trials to optimize the kinematic morphology of a bilateral system. A few tasks were carefully chosen to cover the variability of small scale tasks. For example, the suturing task needs more orientational dexterity compared to the soldering task, which needs a larger workspace and more positional dexterity. The chosen tasks along with their high-level characteristics are listed below (refer to Figure \ref{fig:EM_data}). The choices of task and their characteristics are listed below: 
\begin{enumerate}
    \item \textbf{Pick and Place, Dynamic Bandwidth:}
    e.g. Soldering. Soldering involves picking IC components, placing them on the PCB and finally soldering them to the board. The main characteristics of the task are precision, workspace, accuracy and stiffness control of the system.  
    \item \textbf{Orientation Dexterity:}
    e.g. Suturing. Suturing requires a smaller work-space but regular modulation of the orientation over a broader range.
    \item \textbf{External Wrench Generation:}
    e.g. Knife Cutting. Cutting along a surface path requires positional and wrench generation to overcome contact frictional forces. 
    \item \textbf{Position Dexterity:}
    e.g. Small scale path tracking, incision paths
\end{enumerate}
\subsection{Kinematic Data Collection and Abstraction}
Raw kinematic data was collected through an EM motion-tracking setup that is based on a fixed reference frame. In the collection phase, human expertise on each above individual task was given two manual tools (e.g. forceps with a needle). The base motion of each tool (shown in Figure \ref{fig:tool_setup}) was tracked using an EM tracker probe. The motion of the tip was obtained using a projection matrix as described in \cite{Wijayarathne2017}. 
\par
\begin{figure}[h!] 
\centering
\includegraphics[width=0.85\linewidth]{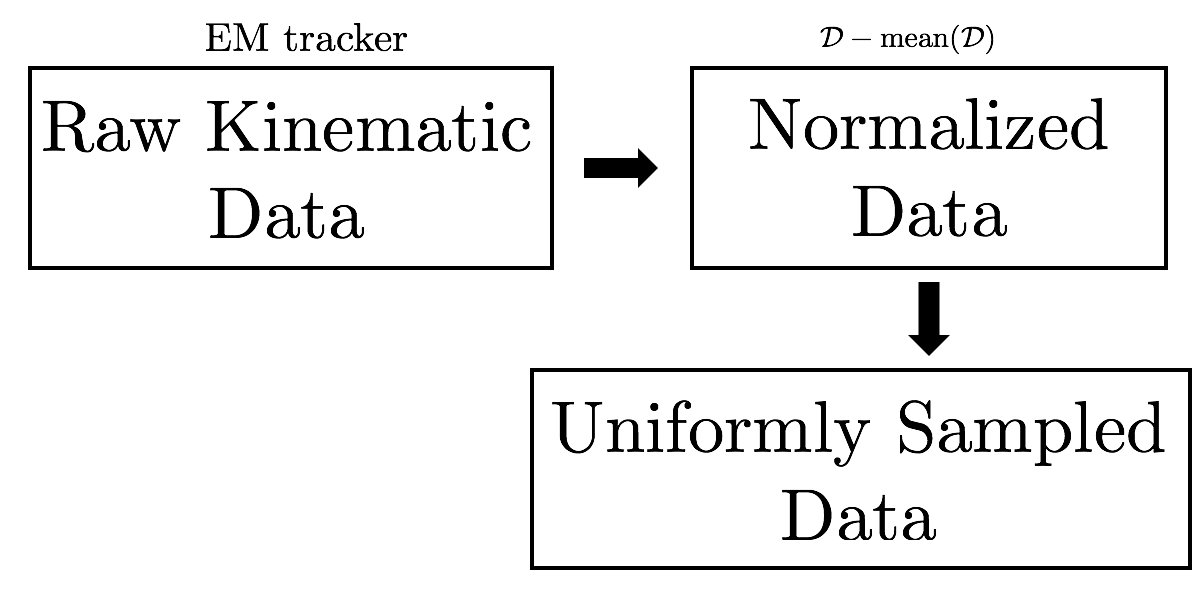}
\caption{Flow of the data pre-processing. Raw data is processed to generate uniformly distributed data}
\label{fig:dataProcess}
\end{figure}

 In performing small scale manipulation tasks, we pay attention to continuous local variations of the motion envelope. This is due to prior global positioning of the body pose. Thus, data was post-processed to extract local variations in motion as shown in Figure \ref{fig:dataProcess}. Normalized data (with local variation) is clustered and re-sampled to generate a uniform distribution. This is important since the kinematic data collected tends to be non-uniform. (e.g., data tends to concentrate in some spatial domains). 

\begin{figure}[h!] 
\centering
\includegraphics[width=0.8\linewidth]{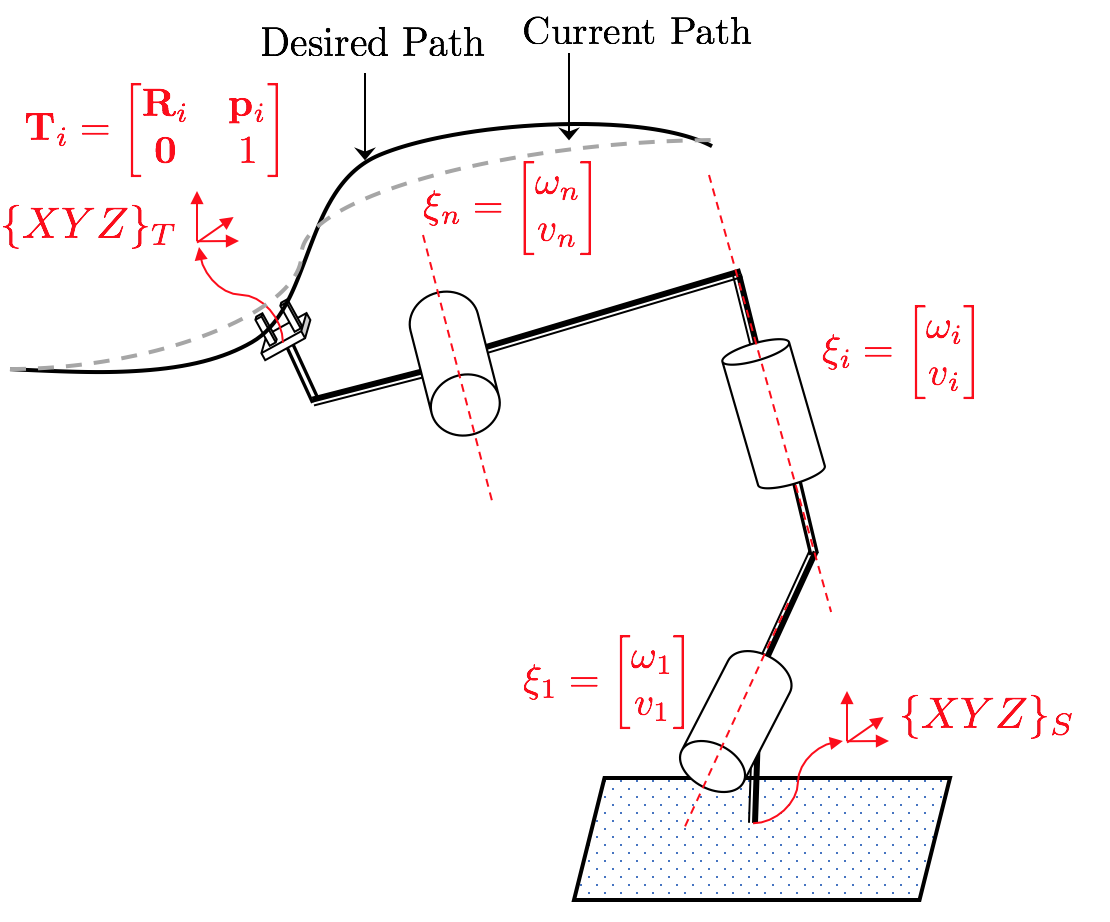}
\caption{Serial Chain Manipulator with revolute joints with product of exponential notation}
\label{fig:serialChain}
\vspace{-15pt}
\end{figure}
\subsection{Preliminaries of Multibody Kinematics}
\subsubsection{Product of Exponential}
The methodology used in this paper uses the basic product of exponential methods for robot kinematic analysis. Compared to the \textbf{Denavit-Hartenberg} ($\mathbf{DH}$) parameters, \textbf{product of exponentials} (\textbf{PoE}) provides a robust way to compute differential inverse kinematic solutions (e.g. positions and orientations). Product of exponentials defines twist axes in the spatial frame rather than relative to the prior frame as seen in the Denavit-Hartenberg convention. As a result, it offers more intuitive solutions in the design morphology optimization that will be shown in the subsequent sections.
\subsubsection{Forward Kinematics}
Given the joint positions, forward kinematics are defined in the product of exponential formulation as below\cite{Murray2017AManipulation}. 
\begin{equation}
\mathbf{T}_{st} = e^{\boldsymbol{\hat{\xi}}_1\theta_1} \hspace{2pt} e^{\boldsymbol{\hat{\xi}}_2\theta_2} \dots e^{\boldsymbol{\hat{\xi}}_i\theta_i} \dots e^{\boldsymbol{\hat{\xi}}_n\theta_n} \hspace{2pt} \mathbf{\!M} \in SE(3)
\end{equation}
$\boldsymbol{\hat{\xi}}_i$ is the twist and represented by $\boldsymbol{\hat{\xi}}_i = \begin{bmatrix}
\hat{\boldsymbol{\omega}}_i & \mathbf{v}_i \\ 0 & 0 \end{bmatrix} \in se(3)$ \footnote{$\hat{\boldsymbol{\omega}} \in so(3)$ is the skew symmetric matrix formed by angular twist coordinates}. $\mathbf{\!M} \in SE(3)$ is home configuration and $\theta_i$'s are the generalized coordinates of the joint positions. 
\subsubsection{Differential Inverse Kinematics}
In general, for a well defined serial manipulator, workspace volume and reachability are known a priori. However, for a manipulator where the morphology is iteratively optimized through a global optimizer (e.g. simulated annealing), inverse solver should be capable of being robust to ill-posed manipulator morphologies throughout the optimization routine. Furthermore, an appropriate IK solver would generate optimized solutions for each iteration where the cost is evaluated. Algorithm \ref{diffINV} shows the modified first order differential IK solver adapted from \cite{Lynch2017MODERNCONTROL,Murray2017AManipulation}.
\begin{algorithm}[!h]
  \caption{Robust Differential Kinematic Solutions }\label{diffINV}
  \begin{algorithmic}[1]
    \Procedure{Differential Inverse Solution}{}
      \State $\text{Initial configuration}\gets \text{random seed}$
      \State $\text{Desired pose}\gets \text{given by data}$
      \State \text{e} = 1 \Comment{initialize error}
      \State \text{$\alpha$} = 0 \Comment{step size}
      \While{$\text{norm}(V) \not= \epsilon$ or $k \leq max$}
        
        \State $\mathbf{T_{sb}}\gets  \mathcal{FK}(\boldsymbol{\theta})$ \Comment{current pose}
        \State $\!V \gets [\text{Ad}_{\mathbf{T_{sb}}}]\log(\mathbf{T^{-1}_{sb}}\mathbf{T})^{\vee}$ \Comment{error in twist coordinates}
        \State $\mathbf{J} \gets \text{Spatial Jacobian}(\boldsymbol{\theta})$ 
        \If  {$\text{rcond}[\mathbf{J}\mathbf{J}^T]<0.001$} \Comment{check singularities}
        \State $\mathbf{J}^{\dag} \gets \mathbf{J}\mathbf{W}^{-1}\mathbf{J}^{T}(\mathbf{J}\mathbf{W}^{-1}\mathbf{J}^T) + 0.001\mathbf{I}_n$
        \Else
        \State $\mathbf{J}^{\dag} \gets \mathbf{J}\mathbf{W}^{-1}\mathbf{J}^{T}(\mathbf{J}\mathbf{W}^{-1}\mathbf{J}^T)$
        \EndIf
        \State $\boldsymbol{\theta} \gets \boldsymbol{\theta} + \alpha\mathbf{J}^\dag\hspace{2pt}\!V$ 
      \EndWhile\label{euclidendwhile}
      \State \textbf{return} $\boldsymbol{\theta}$
    \EndProcedure
  \end{algorithmic}
\end{algorithm}
\vspace{-10pt}
\section{Design optimization}\label{sec:designOpt}
\subsection{Robot Morphology Optimization}
In Section \ref{sec:fundamentals}, the fundamentals needed for the formulation of the optimization is given. In this section, we use the fundamentals to formulate the design problem. 
\subsubsection{Geometric optimization for Morphology}
The model of a robot is dependant on its geometrical parameters as well the inertial parameters. For fabrication and physical feasibility, optimization of geometrical parameters is easier to achieve, whereas inertial parameters are dependent on actuator variants (e.g., brushed, brush-less DC, gearing, stiffness, etc.), fabrication material, and transmission mechanisms. In this paper, we consider a task-driven manipulator optimization where \textbf{\underline{local variations of kinematic data}} from the tasks are used to optimize each of the symmetric manipulators of the system. 
\subsubsection{Gradient-free optimization for the above data metrics}
Since the cost function for the manipulator optimization is non-linear, gradient-based optimization techniques are not well suited. Hence we used simulated annealing to retrieve the optimal parameters. With an appropriate number of iterations and uniform sampling over the design parameters space, it can be assumed to converge to the optimal parameter set. Algorithm
\ref{gradientOptim} shows the procedural pseudo-code of the steps used.
\subsubsection{Design Parameters}
\begin{figure}[t!] 
\centering
\includegraphics[width=0.9\linewidth]{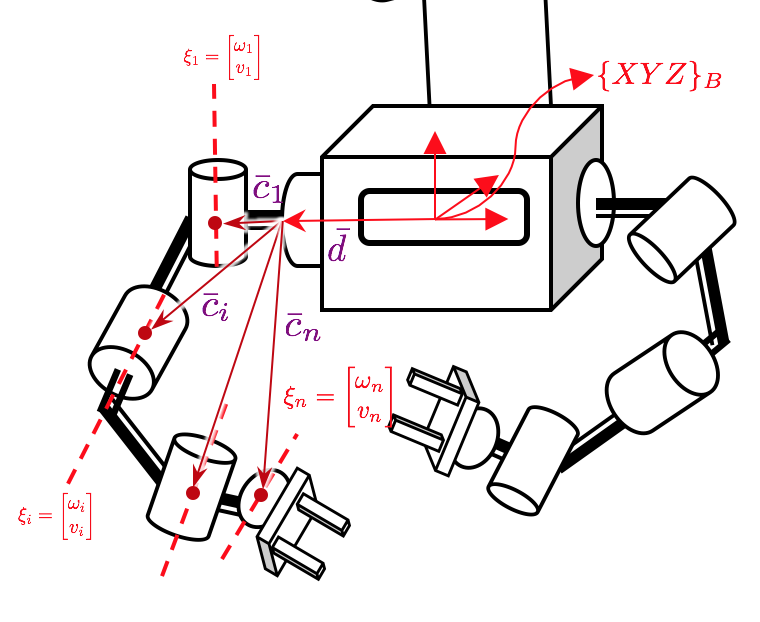}
\caption{Illustration of the Design parameters in the dual manipulator system}
\label{fig:diff_control}
\end{figure}

In the optimization framework, we assume both the arms are symmetric\footnote{This is to facilitate, especially in teleoperation, right and left hand dominancy}.  Thus, optimization is done with symmetric arms on bi-manual tasks. Figure \ref{fig:diff_control} visualizes the design parameters that were used. Here, $\bar{\mathbf{c}}_i$ is the distance to each joint frame from the base frame $\{XYZ\}_B$ and $\vg \xi_i$ is the twist at $\bar{\mathbf{c}}_i$. $\bar{\mathbf{d}}$ is the distance vector to the first joint from the base frame. Moreover, we impose a design structural constraint ($\bar{\mathbf{c}}_5 = \bar{\mathbf{c}}_6 = \bar{\mathbf{c}}_7$) for the wrist such that last three axes intersect ($\vg \xi_5, \vg \xi_6, \vg \xi_7$). This is to ensure fabrication feasibility and prior morphological initialization.
\par
For certain bilateral tasks, the center distance $\bar{\mathbf{d}}$ is important. For instance, in tasks with a small work-space volume (e.g. soldering), the center distance between two manipulators could be small. This will allow inter-crossing between two arms which is preferable in some tasks. However, the wider the center distance, the more convenient it becomes to do bilateral manipulation of larger objects. 
\subsubsection{Cost Function for Design Optimization}
Since the application focus of this work is to get a design candidate that can achieve pose variations, we design the cost function to achieve the desired poses and have a well-connected workspace. We define the cost function for design as:
\begin{multline}
    \mathcal{C} = \sum_{i = 1}^N \hspace{2pt}\delta \v p^T\v W_1 \delta \v p + \v W_2(\bar{\v q}_c[i]-\bar{\v q}_c[i-1]) 
    \label{eq:prime_objective}
\end{multline}
where, $\delta \v p\footnote{$\v T_{sb}$ is the current pose and $\v T$ is the desired pose} = [\text{Ad}_{\mathbf{T_{sb}}}]\log(\mathbf{T^{-1}_{sb}}\mathbf{T})^{\vee}$, $\v W_1$ and $\v W_2$ are weighting matrices. To ensure a well-connected workspace, the error term  $(\bar{\v q}_c[i]-\bar{\v q}_c[i-1])$  between last and current joint vectors needs to be minimized.
\begin{algorithm}[!h]
  \caption{Kinematic Optimization Routine}\label{gradientOptim}
  \begin{algorithmic}[1]
    \Procedure{Morphology Optimization}{}
      \State $\text{Robot Morphology}\gets \text{twist and distance to axis}$
      \State \text{err} = 1 \Comment{initialize error}
      \State \text{i} = 0 \Comment{iteration}
      \State $\mathcal{R}_0$ = \text{Robot Morphology} \Comment{initial robot}
      \State $\mathcal{D}[i]\gets  \text{post-processed local pose data, normalized}$
      \\
      \While{$e \not= \epsilon$ or $i \leq \text{max}_i$}
        \State $\mathcal{D}[i] \gets  \mathcal{D}[i] + [\bar{\mathbf{c}}_5, \text{R}_5]$ \Comment{adds current wrist pose to point data cloud}  \label{alg1:data} 
        \\
        \State $\mathcal{C}[i]\gets \text{IK Cost}(\mathcal{D}[i])$
        \State $\text{Temp.} = \mathcal{F}[i]$  \Comment{temp. exponential decay}
        \State $\mathcal{R}_{i+1} \gets \mathcal{G}{(P[\mathcal{R}[i]], \text{Temp.})}$ \Comment{new candidate (constrained)}
        \State $\text{Update} \quad P[\mathcal{R}[i], \mathcal{C}[i]]$ \Comment{update the distribution}
      \EndWhile\label{euclidendwhile}
      \\
      \State \textbf{return} $\mathcal{R}_n$ \Comment{Optimized Robot}
    \EndProcedure
  \end{algorithmic}
\end{algorithm}
\subsection{Design Optimization}
We used simulated annealing (SA) for the optimization of the post-processed data. SA was run with uniform re-sampling to select new candidates. At each iteration, the motion data set with the local cloud was added to the current wrist pose ($\bar{\mathbf{c}}_5$) [Line \ref{alg1:data} of Algorithm 1]. This is because the data is normalized and the optimal dexterous pose of the wrist from the is not known beforehand.
The optimization process yields an optimal set of design parameters after an appropriate number of iterations with no further notable improvement. 
Data processing and optimization were done with the aid of MATLAB$^\copyright$ 2019A. The global optimization toolbox provided by MathWorks$^\copyright$ was used in this work.
\section{Design Informed Motion Optimization}
In Section \ref{sec:designOpt}, the optimal pose (the distance vector and orientation from the base, $\bar{\mathbf{c}}_5$) in which the robot is most dexterous is calculated\footnote{Data is local variations around a pose relative to the base}. Since this information is known a priori, it is possible to infuse this information to robot motion generation. 
\subsection{Spatial Jacobian of the Manipulator Tree}
 We leverage the redundancy of the system in motion generation along with the design limitations and prior information (e.g., where the most dexterous location is relative to the base). We propose a kinematic differential solver that will use the known information to optimize the configuration of the overall system. 
\begin{equation}
    \bar{\mathbf{J}} = 
    \begin{bmatrix}
    \begin{bmatrix} \mathbf{J}_K & \text{Ad}_{\mathbf{T_{sk}}\mathbf{J}_L}\end{bmatrix} & \mathbf{0} \\
    \mathbf{0} & \begin{bmatrix} \mathbf{J}_K & \text{Ad}_{\mathbf{T_{sk}}\mathbf{J}_R}\end{bmatrix}
    \end{bmatrix}
\end{equation}
$\bar{\mathbf{J}}$ is the overall spatial jacobian of the system where, $\mathbf{J}_K$ indicates the spatial jacobian of the global positioner system (industrial manipulator), $\mathbf{J}_L$ and $\mathbf{J}_R$ are the left and right mini manipulators. In solving the differential motion, inverse of the $\bar{\mathbf{J}}$ should be defined. 
\subsection{Generalized Inverse}
In comparison to single-arm serial manipulators, dual-arm systems have a higher order of redundancy. Thus, in order to get inverse kinematic solutions\cite{SicilianoB.SciaviccoL.VillaniL.OrioloRobotAnalysis}, differential kinematic solutions are often used. Kinematic redundancy of the system could be used to satisfy a secondary objective function in addition to satisfying the desired pose ($\in SE(3)$). The inverse of the Jacobian matrix for a redundant system is not unique and subject to satisfy a user-defined cost function. A commonly used inverse is the Moore–Penrose inverse where $||\dot{\vg \theta}||=\sqrt{\dot{\vg \theta}^T\dot{\vg \theta}}$ is minimized. In general, the formulation could be written as:
\begin{align*}
     & \underset{\vg \theta}{\text{min}} \ \frac{1}{2} \dot{\vg \theta}^T\v M(\vg \theta)\dot{\vg \theta} + \nabla \mathcal{H}(\vg \theta)^T\vg \dot{\vg \theta} \\
        & \text{subject to} \ \mathcal{V}_d = \bar{\v J}(\vg \theta)\dot{\vg \theta}, \ \underline{\vg \theta} \leq \vg \theta \leq \bar{\vg \theta}
\end{align*}
where, $\mathcal{H}$ is a user defined artificial potential field (described in Section \ref{sec:null_space_cost}). $\v M(\vg \theta)$ is the inertial matrix of the overall system and $\mathcal{V}_d$ is the desired spatial velocity. The solution to this optimization could be written as\footnote{full derivation could be found on Appendix D of \cite{Lynch2017MODERNCONTROL}}:
\begin{align}
    \dot{\vg \theta} = \v G\mathcal{V}_d + (\v I - \v G\bar{\v J})\v M^{-1}\nabla \mathcal{H} \\
    \v G = \v M^{-1}\bar{\v J}^T(\bar{\v J}\v M^{-1}\bar{\v J}^T)^{-1}
\end{align}
where, $\nabla \mathcal{H}$ is the partial derivative with respect to each joint coordinate.
\subsection{Null Space Cost Function}
\label{sec:null_space_cost}
We define a secondary cost function\footnote{Primary function is to get to the pose commanded by the user} where we can leverage the redundancy and use the prior knowledge about manipulator dexterity in the motion planner as shown in the Figure \ref{fig:diff_control}. The function can be expressed as:
\begin{multline}
    \mathcal{H} = \sum_{i = L, R} (\bar{\v w}_i-(\bar{\v r}_i+\bar{\v t}_i))^T\v Q_i (\bar{\v w}_i-(\bar{\v r}_i+\bar{\v t}_i))  
    \label{eq:null_space_function}
\end{multline}
where, $\v Q_i$ is the weighting matrix, $\bar{\v r}_i$ is the distance to the current end-effector location of each arm, $\bar{\v w}_i$ is the desired pose of the desired workspace pose and $\bar{\v t}_i$ is the tool geometry of the left and right arms respectively. All vectors are defined relative to the $\{XYZ\}_B$ in the global world frame.
\begin{figure}[h!] 
\centering
\includegraphics[width=0.8\linewidth]{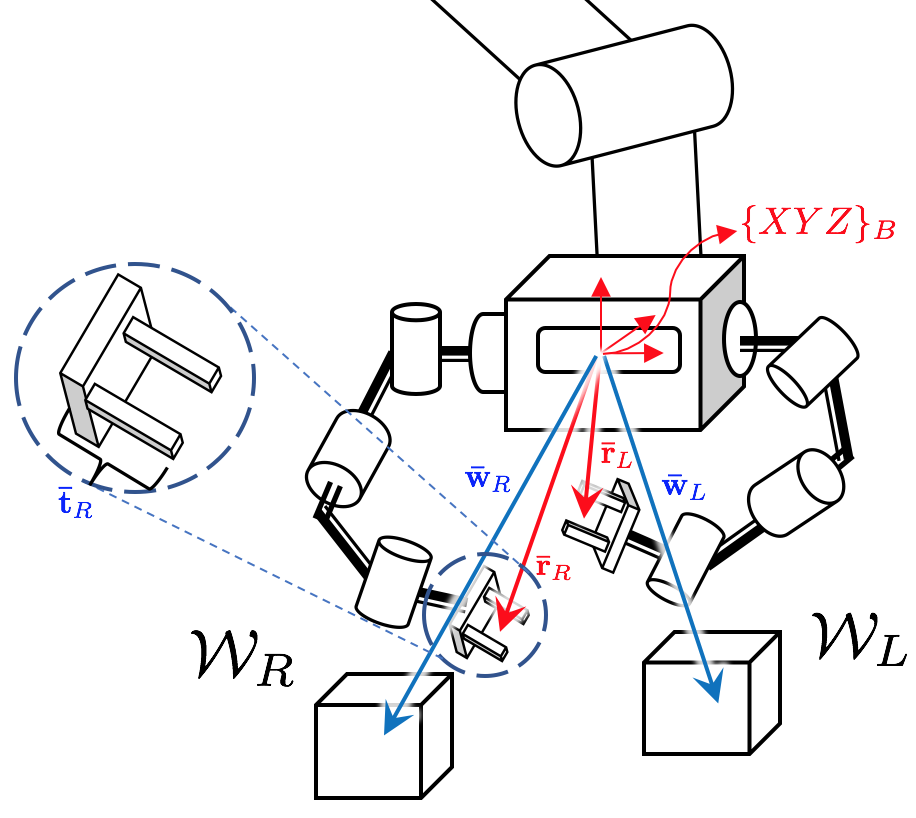}
\caption{Bilateral system with desired workspace poses and current pose of the manipulator system}
\label{fig:design_parameters}
\end{figure}

\begin{figure}[h!] 
\centering
\includegraphics[width=0.8\linewidth]{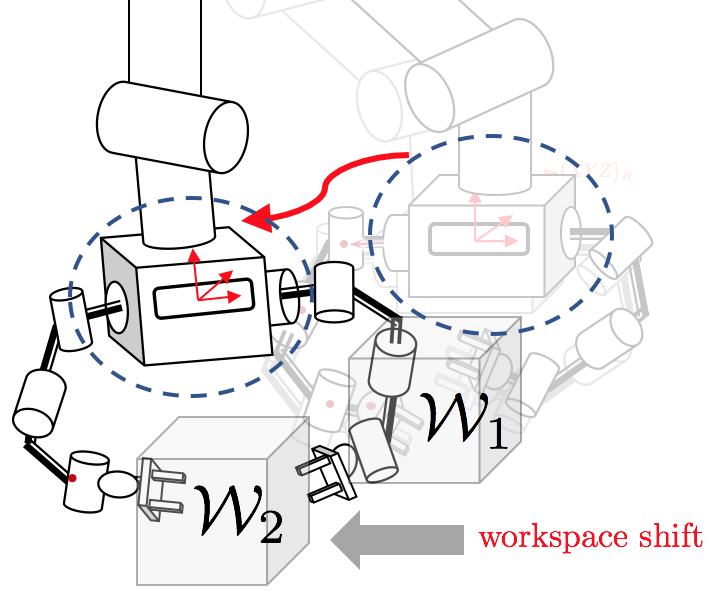}
\caption{Motion of the global to maintain the dexterity through task transitions ($\mathcal{W}_1 \rightarrow \mathcal{W}_2$).}
\label{fig:motion_task}
\end{figure}

\begin{figure*}[t!] 
\centering
\includegraphics[width=\textwidth,height=9cm]{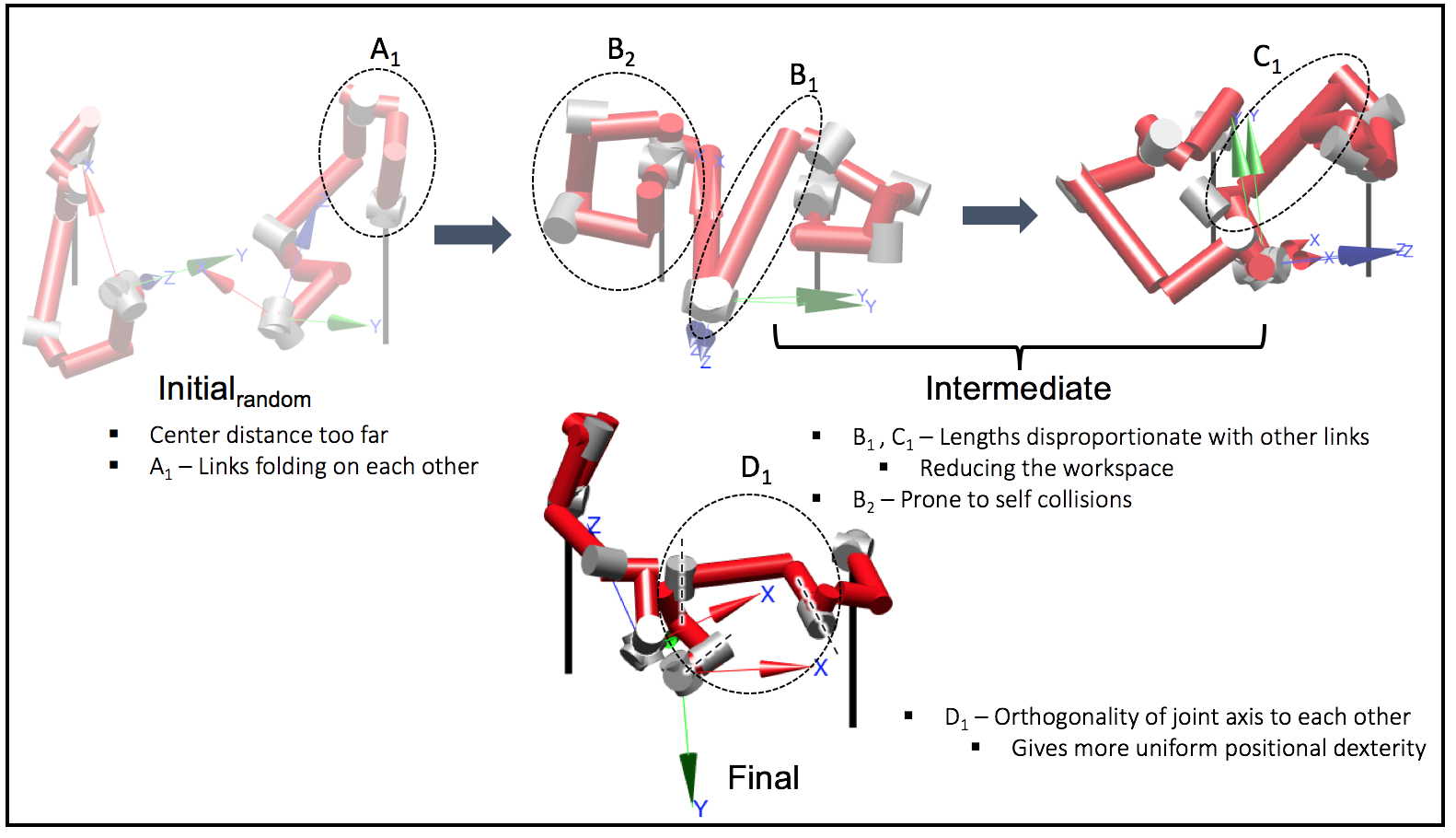}
\caption{Evolution of the design parameters over iterations. A few intermediate states.}
\label{fig:design_iterations}
\vspace{-15pt}
\end{figure*}
\subsection{Adaptation to Task through Inertial Matrix $\v M(\vg \theta)$}
In general, $\v M(\vg \theta)$ is defined to be the mass matrix of the entire manipulator tree. We use the mass matrix with a scaling parameter to be adaptive to the task in hand. For instance, if the desired and predetermined dexterous work-space of the system is at a far distance, the cost of $\mathcal{H}$ is made dominant. In such case, motion to the new work-space is achieved by the movement of the base. We modulate the scaling of the mass matrix of each segment via:
\begin{equation}
    \v M(\vg \theta) = diag\big(
    \bar{\v M}_K(\vg \theta_K) \quad \bar{\v M}_L(\vg \theta_L) \quad \bar{\v M}_R(\vg \theta_R)\big)
\end{equation}
where, $\v M_K(\vg \theta_K)$, $\v M_L(\vg \theta_L)$, and $\v M_R(\vg \theta_R)$ are the mass matrices of the base manipulator, left and right arms respectively. Each of the mass matrices were scaled as desired by the user and the task in hand. We used the following continuous switching law where $\beta_i$ is the threshold of the activation of each manipulator segment based on the left and right end-effector poses relative to the workspace. This can be defined as a binary switching law if desired.
\begin{equation}
    \bar{\v M}_i(\vg \theta_i) = \big( \tanh{(|\v w_i - \v r_i|_2 - \beta_i)} + 1.5\big)\v M_i(\vg \theta_i)
    \label{eq:m_update}
\end{equation}
Intuitively, the reasoning for the above-chosen scaling parameters is as follows. The joint configuration is optimized continuously while following the desired Cartesian trajectory as commanded by the user. Secondary function for null space optimization is given by Eq. \ref{eq:null_space_function} and Eq. \ref{eq:m_update} updates the weighting matrices of each manipulator segment of the kinematic tree. This facilitates the user to manipulate the relative motion of kinematic segments while leveraging different bandwidth capabilities needed for the task (for example, certain tasks need high bandwidths and manipulator segments with low inertia are better suited for actuation).\footnote{We used only the base manipulator with a activation function here} 


\subsection{Kinematic Differential Motion Control}
Once the optimum design morphology was achieved, the optimal distance vector to the dexterous location is known ($\mathbf{\bar{r}} = \bar{\mathbf{c}}_5$). It is to be noted that the obtained dexterity is only valid when $\mathbf{\bar{r}}$ is maintained throughout the task completion. This objective is achieved through the continuous motion of the base using a weighting matrix $\mathbf{\v M(\vg \theta)}$. \par 
Moreover, this motion can undesirable in cases where the system is teleoperated due to the change of the view perspective. In such a scenario, the weights corresponding to the base could be made larger. On the other hand, in autonomous tasks, continuous optimization of the configuration can improve task safety, performance, and dexterity.

\subsection{Conversion of PoE to DH}
Once the optimum morphology of the system is found, we use the method used by \cite{Wu2017AnRobots} to convert obtained PoE parameters to DH (Denavit-Hartenberg). This conversion was used as there isn't a visualization tool for robot simulation with PoE parameters. The visualization was done with the aid of Peter Corke's robotics toolbox \cite{Corke2017RoboticsRevised} with standard DH parameters (converted).
\section{Kinematic Performance Evaluation}
\label{sec:dexterity}
In order to evaluate the performance of the optimized design and kinematic control scheme, standard performance metrics were used. The global metric to asses the performance is can be written as:
\begin{multline}
    \mathcal{E} = \underbrace{\sum_{i = 1}^{N}   \Big(\frac{\max{\lambda[\mathbf{J}_i\mathbf{J}_i^T]}}{\min{\lambda[\mathbf{J}_i\mathbf{J}_i^T]}} - 1\Big)  + \det{\mathbf{J}_i\mathbf{J}_i^T}}_A \\ + \underbrace{(\v q - \bar{\v q})^2 + (\v q - \underline{\v q})^2}_B
\end{multline}
where $\v J_i$ is the spatial jacobian and $\bar{\v q}$  and $\underline{\v q}$ are the joint limits. $A$ and $B$ are the terms that corresponds to dexterity measure and joint limit constraint respectively. 

\begin{figure}[b!] 
\centering
\includegraphics[width=0.9\linewidth]{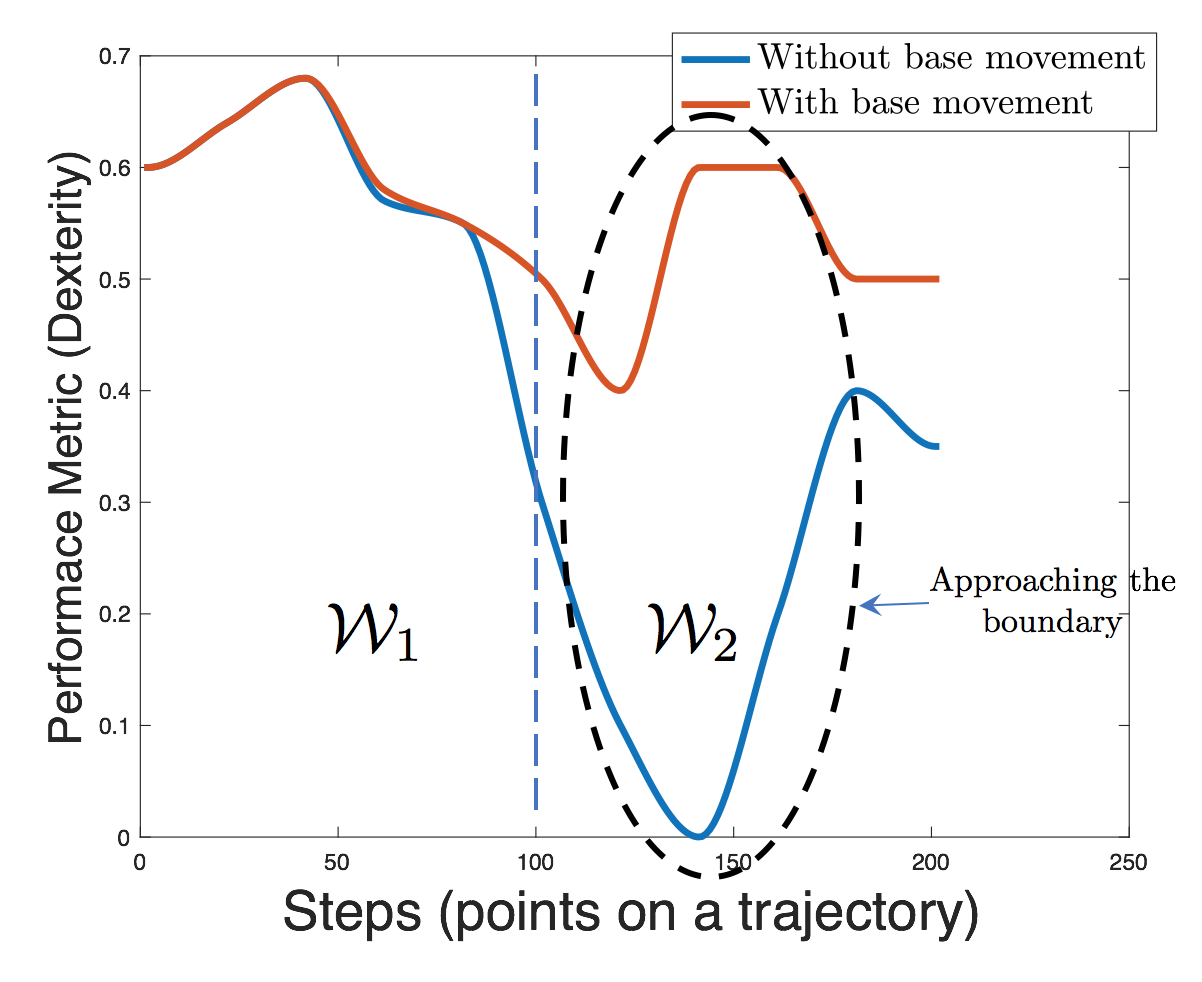}
\caption{Performance comparison with and without design informed motion optimization}
\label{fig:performance_design_motion}
\end{figure}

\section{Analysis}
\begin{figure}[b!] 
\centering
\includegraphics[width=1\linewidth]{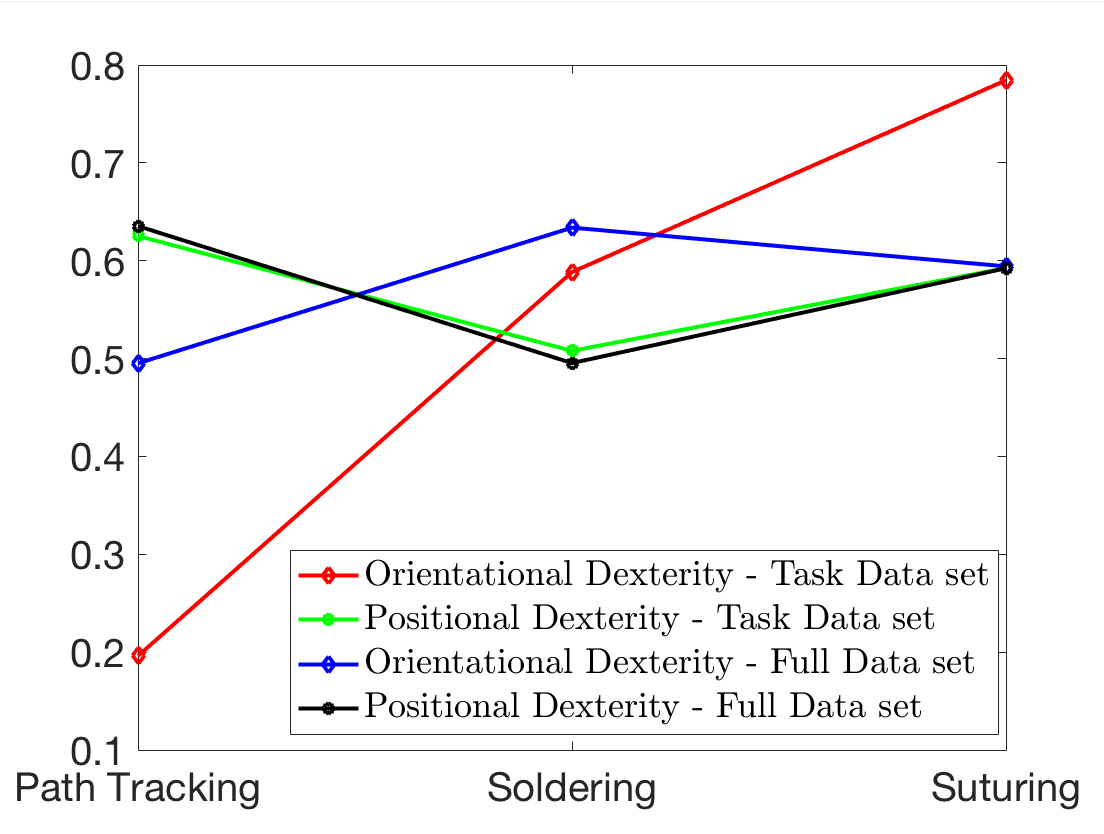}
\caption{Performance comparison over different tasks. Compared over individual task data and on full data set}
\label{fig:performance_tasks}
\end{figure}

We explored the possibility of optimizing the manipulator geometry and then using the optimal design parameters in the kinematic motion generation. To evaluate the performance gain, we used two methods. \begin{inparaenum}[1)]
\item Kinematic dexterity comparison over optimizing each individual task vs all tasks (shown in Figure \ref{fig:performance_tasks}). 
\item Kinematic dexterity  throughout the task with and without design informed motion (Figure \ref{fig:performance_design_motion})\end{inparaenum}  \par 
We used a data-driven approach to optimize morphological design parameters for a manipulator arm with seven DOFs (one of the symmetric arms). As discussed in Section \ref{sec:designOpt}, we had imposed morphological constraints to facilitate physical feasibility. We show that the optimal design obtained by utilizing pose data throughout all tasks would result in a more versatile design morphology. Figure \ref{fig:performance_tasks} shows the performance comparison over three different tasks that represent the motion characteristics. The performance metric (from Section \ref{sec:dexterity}) is normalized such that orientational and positional metrics could be represented on the same axis. \par
To show the successful incorporation of design information to motion, we chose two work-spaces adjacent to each other (as shown in Figure \ref{fig:motion_task}). We chose target positions in work-space regions $\mathcal{W}_1$ and $\mathcal{W}_2$. The goal was to switch between these workspaces with and without design informed motion optimization. Figure \ref{fig:performance_design_motion} shows dexterity evaluation during the transition. It can be observed that dexterity throughout the duration is more uniform when designed informed motion generation is used. This is due to motion compensation of the base throughout the task to keep the dexterity of the bilateral system optimal. However, the motion of the base is not desired in cases where manipulation tasks require high bandwidth and precise performance. In such scenario, motion of each segment can be  modulated by defining custom thresholds in Eq. \ref{eq:m_update}.  This approach would facilitate smooth transitioning and near uniform dexterity throughout task manipulation.


\section{Conclusion and Future Work}
In this paper, we explored the possibility of optimizing robot morphology for positional and orientational dexterity. Data for optimization was obtained through kinematic motion capture (through an EM tracker) of a human subject performing a few representative bimanual small scale tasks (e.g, soldering, suturing, path tracking). The information gained out of morphology optimization was employed to optimize the differential motion. The design obtained by using the complete motion data set is uniformly dexterous over all tasks. Moreover, it was shown that by using design informed motion generation, dexterity was increased by $\sim 10\%$ throughout the task.
\par
Future work would include real-time implementation and validation of these results as well as implementation in autonomous dual manipulation tasks.

\addtolength{\textheight}{-12cm}   





\bibliographystyle{ieeetr}
\bibliography{refs}

\begin{thebibliography}{10}

\bibitem{Dietrich2020HierarchicalRobots}
A.~Dietrich and C.~Ott, ``{Hierarchical Impedance-Based Tracking Control of
  Kinematically Redundant Robots},'' {\em IEEE Transactions on Robotics},
  vol.~36, no.~1, pp.~204--221, 2020.

\bibitem{Smith2012DualSurvey}
C.~Smith, Y.~Karayiannidis, L.~Nalpantidis, X.~Gratal, P.~Qi, D.~V.
  Dimarogonas, and D.~Kragic, ``{Dual arm manipulation—A survey},'' {\em
  Robotics and Autonomous Systems}, vol.~60, pp.~1340--1353, 10 2012.

\bibitem{Park2016Dual-armEvaluation}
H.~A. Park and C.~S.~G. Lee, ``{Dual-arm coordinated-motion task specification
  and performance evaluation},'' in {\em 2016 IEEE/RSJ International Conference
  on Intelligent Robots and Systems (IROS)}, pp.~929--936, IEEE, 10 2016.

\bibitem{Paden2010DesignManipulators}
B.~Paden, ``{Design of 6R Manipulators},'' pp.~43--61, 2010.

\bibitem{Ranjbaran1995TheIsotropy}
F.~Ranjbaran, J.~Angeles, M.~A. Gonz{\'{a}}lez-Palacios, and R.~V. Patel,
  ``{The mechanical design of a seven-axes manipulator with kinematic
  isotropy},'' {\em Journal of Intelligent {\&} Robotic Systems}, vol.~14,
  no.~1, pp.~21--41, 1995.

\bibitem{Vijaykumar1986}
R.~Vijaykumar, K.~Waldron, and M.~Tsai, ``{Geometric Optimization of Serial
  Chain Manipulator Structures for Working Volume and Dexterity},'' {\em The
  International Journal of Robotics Research}, vol.~5, pp.~91--103, 6 1986.

\bibitem{Rodriguez2013EffectorActuation}
A.~Rodriguez and M.~T. Mason, ``{Effector form design for 1DOF planar
  actuation},'' {\em Proceedings - IEEE International Conference on Robotics
  and Automation}, pp.~349--356, 2013.

\bibitem{Shirafuji2019KinematicKinematicsb}
S.~Shirafuji and J.~Ota, ``{Kinematic Synthesis of a Serial Robotic Manipulator
  by Using Generalized Differential Inverse Kinematics},'' {\em IEEE
  Transactions on Robotics}, vol.~1, pp.~1--8, 2019.

\bibitem{Shiakolas2002}
P.~Shiakolas, D.~Koladiya, and J.~Kebrle, ``{Optimum Robot Design Based on Task
  Specifications Using Evolutionary Techniques and Kinematic, Dynamic, and
  Structural Constraints},'' {\em Inverse Problems in Engineering}, vol.~10,
  pp.~359--375, 1 2002.

\bibitem{Rao1989OptimizationSurwef}
S.~S. Rao and P.~K. Bhatti, ``{Optimization in the design and control of
  robotic manipulators: A surwef},'' tech. rep., 1989.

\bibitem{Webster2018KinematicPlanning}
R.~J. Webster, R.~Alterovitz, P.~L. Anderson, C.~Baykal, A.~W. Mahoney,
  C.~Bowen, A.~Kuntz, and F.~Maldonado, ``{Kinematic Design Optimization of a
  Parallel Surgical Robot to Maximize Anatomical Visibility via Motion
  Planning},'' {\em 2018 IEEE International Conference on Robotics and
  Automation (ICRA)}, pp.~926--933, 2018.

\bibitem{Computational-Co-Optimization-of-Design-Parameters-and-Motion-Trajectories-for-Robotic-Systems-Paper.pdf}
``{Computational-Co-Optimization-of-Design-Parameters-and-Motion-Trajectories-for-Robotic-Systems-Paper.pdf}.''

\bibitem{Gosselin1991AManipulators}
C.~Gosselin and J.~Angeles, ``{A Global Performance Index for the Kinematic
  Optimization of Robotic Manipulators},'' {\em Journal of Mechanical Design},
  vol.~113, p.~220, 9 1991.

\bibitem{Hammond2009MorphologicalMeasures}
F.~L. Hammond and K.~Shimada, ``{Morphological design optimization of
  kinematically redundant manipulators using weighted isotropy measures},''
  {\em Proceedings - IEEE International Conference on Robotics and Automation},
  pp.~2931--2938, 2009.

\bibitem{Asada2009ADesign}
H.~Asada, ``{A Geometrical Representation of Manipulator Dynamics and Its
  Application to Arm Design},'' {\em Journal of Dynamic Systems, Measurement,
  and Control}, vol.~105, no.~3, p.~131, 2009.

\bibitem{KhatibOptimizationManipulators}
O.~Khatib and A.~Bowling, ``{Optimization of the Inertial and Acceleration
  Characteristics of Manipulators},'' no.~1.

\bibitem{Shirafuji2019KinematicKinematics}
S.~Shirafuji and J.~Ota, ``{Kinematic Synthesis of a Serial Robotic Manipulator
  by Using Generalized Differential Inverse Kinematics},'' {\em IEEE
  Transactions on Robotics}, vol.~1, pp.~1--8, 2019.

\bibitem{Hannaford2013Raven-II:Research}
B.~Hannaford, J.~Rosen, D.~W. Friedman, H.~King, P.~Roan, L.~Cheng, D.~Glozman,
  J.~Ma, S.~N. Kosari, and L.~White, ``{Raven-II: An open platform for surgical
  robotics research},'' {\em IEEE Transactions on Biomedical Engineering},
  vol.~60, no.~4, 2013.

\bibitem{GuthartTheApplication}
G.~Guthart and J.~Salisbury, ``{The Intuitive/sup TM/ telesurgery system:
  overview and application},'' in {\em Proceedings 2000 ICRA. Millennium
  Conference. IEEE International Conference on Robotics and Automation.
  Symposia Proceedings (Cat. No.00CH37065)}, vol.~1, pp.~618--621, IEEE.

\bibitem{Lee2015RedundancyExperimentsb}
J.~Lee and P.~H. Chang, ``{Redundancy resolution for dual-arm robots inspired
  by human asymmetric bimanual action: Formulation and experiments},'' in {\em
  2015 IEEE International Conference on Robotics and Automation (ICRA)},
  pp.~6058--6065, IEEE, 5 2015.

\bibitem{Rakita2019SharedManipulation}
D.~Rakita, B.~Mutlu, M.~Gleicher, and L.~M. Hiatt, ``{Shared control–based
  bimanual robot manipulation},'' {\em Science Robotics}, vol.~4, no.~30, 2019.

\bibitem{Wijayarathne2017}
L.~Wijayarathne and F.~L. Hammond, ``{Kinetic Measurement Platform for Open
  Surgical Skill Assessment},'' {\em Journal of Medical Devices}, 2017.

\bibitem{Murray2017AManipulation}
R.~M. Murray, Z.~Li, and S.~Shankar~Sastry, {\em {A mathematical introduction
  to robotic manipulation}}.
\newblock 2017.

\bibitem{Lynch2017MODERNCONTROL}
K.~M. Lynch, {\em {MODERN ROBOTICS. MECHANICS, PLANNING, AND CONTROL}}.
\newblock No.~May, 2017.

\bibitem{SicilianoB.SciaviccoL.VillaniL.OrioloRobotAnalysis}
S.~L. V. L. O.~G. Siciliano, B., {\em {Robot Modeling, Control and Analysis}}.

\bibitem{Wu2017AnRobots}
L.~Wu, R.~Crawford, and J.~Roberts, ``{An Analytic Approach to Converting POE
  Parameters into D-H Parameters for Serial-Link Robots},'' {\em IEEE Robotics
  and Automation Letters}, vol.~2, no.~4, pp.~2174--2179, 2017.

\bibitem{Corke2017RoboticsRevised}
P.~Corke, {\em {Robotics, vision and control: fundamental algorithms in
  MATLAB{\textregistered} second, completely revised}}, vol.~118.
\newblock Springer, 2017.

\end{thebibliography}

\end{document}